\documentclass[10pt,twocolumn,letterpaper]{article}

\usepackage{cvpr}              
\usepackage{multirow}
\usepackage{makecell}
\usepackage{booktabs}
\usepackage{bm}
\usepackage{amsfonts,amssymb}

\usepackage{soul}

\usepackage[accsupp]{axessibility} 

\usepackage[dvipsnames]{xcolor}

\definecolor{cvprblue}{rgb}{0.21,0.49,0.74}
\usepackage[pagebackref,breaklinks,colorlinks,citecolor=cvprblue]{hyperref}

\newcommand{\issue}[1]{\vspace{0.05in}\noindent \textbf{#1 \hspace{0.1em}}}

\def\paperID{3590} 
\def\confName{CVPR}
\def\confYear{2024}

\title{GOV-NeSF: Generalizable Open-Vocabulary Neural Semantic Fields}

\author{Yunsong Wang \qquad Hanlin Chen \qquad Gim Hee Lee\\
Department of Computer Science, National University of Singapore\\
{\tt\small \{yunsong, hanlin.chen, gimhee.lee\}@comp.nus.edu.sg}\\
{\tt\small \href{https://github.com/wangys16/GOV-NeSF}{\textbf{https://github.com/wangys16/GOV-NeSF}}}}

\begin{document}

\definecolor{scannet}{rgb}{0.886, 0.941, 0.851}
\sethlcolor{scannet}
\definecolor{replica}{rgb}{1, 0.949, 0.8}
\sethlcolor{replica}
\twocolumn[{%
\renewcommand\twocolumn[1][]{#1}%
\maketitle
\begin{center}
    \centering
    \captionsetup{type=figure}
    \includegraphics[width=1\textwidth]{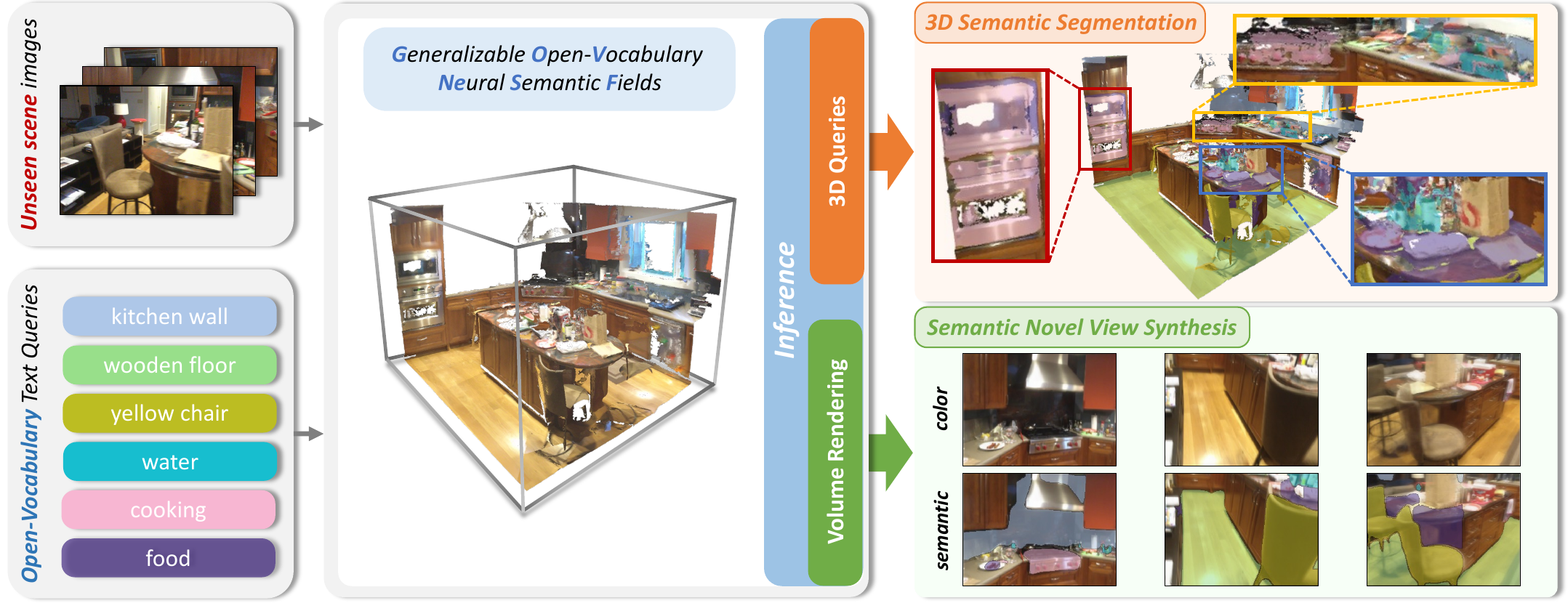}
    \captionof{figure}{\textbf{Overall pipeline of GOV-NeSF.} Given the posed images from any unseen 3D scene, and arbitrary open-vocabulary text queries, our model is capable of both open-vocabulary 3D semantic segmentation and novel view synthesis with 2D semantic segmentation.}
    \label{teaser}
\end{center}%
}]

\begin{abstract}
    \vspace{-1mm}
    Recent advancements in vision-language foundation models have significantly enhanced open-vocabulary 3D scene understanding. However, the generalizability of existing methods is constrained due to their framework designs and their reliance on 3D data. 
    We address this limitation by introducing \textbf{G}eneralizable \textbf{O}pen-\textbf{V}ocabulary \textbf{Ne}ural \textbf{S}emantic \textbf{F}ields (GOV-NeSF), a novel approach offering a generalizable implicit representation of 3D scenes with open-vocabulary semantics. We aggregate the geometry-aware features using a cost volume, and propose a Multi-view Joint Fusion module to aggregate multi-view features through a cross-view attention mechanism, which effectively predicts view-specific blending weights for both colors and open-vocabulary features. Remarkably, our GOV-NeSF exhibits state-of-the-art performance in both 2D and 3D open-vocabulary semantic segmentation, eliminating the need for ground truth semantic labels or depth priors, and effectively generalize across scenes and datasets without fine-tuning.
\end{abstract}

\section{Introduction}

Semantic segmentation for 2D \cite{2d1, 2d2, 2d3, 2d4} and 3D \cite{pointnet, pointnet++, minkowski, 3d1} is a fundamental problem in computer vision with broad applications in fields such as autonomous driving \cite{driving1, driving2}, robotic navigation \cite{navi1}, medical imaging analysis \cite{medical1}, \textit{etc}. Given the input 2D images or 3D data such as point cloud, the model is trained to predict dense semantic labels that are assigned to each pixel or point, respectively.  Conventional approaches for semantic segmentation are limited by a predefined set of classes that can potentially be assigned to a pixel/point, which hinders the 
generalizability of the models across datasets and label sets, resulting in dataset-specific 
models that heavily rely on labels. 

Recently, Vision-Language Models (VLMs) \cite{clip, filip, coca, laion} propose to learn vision-language correlations from web-scale image-text pairs, showing impressive generalizability in zero-shot vision recognition tasks across different datasets. 
Attempts like MaskCLIP \cite{maskclip} and PointCLIP \cite{pointclip} have explored transferring knowledge from 2D VLMs to 3D encoders, benefiting from the robustness and generalizability inherent in 2D VLMs. 
Despite their advantages, these methods typically require pairs of images and point clouds during training, which is largely constrained by the limited availability of 3D data. Additionally, when this knowledge is directly distilled from 2D to 3D contexts, the limited 3D dataset sizes for training may compromise the 
generalizability of the models, which is the key strength of VLMs.

OpenScene \cite{openscene} has achieved notable success in zero-shot and open-vocabulary 3D semantic segmentation by averaging multi-view open-vocabulary features and distilling them into a 3D encoder. 
Despite these achievements, there are several issues with this approach: 1) Averaging multi-view open-vocabulary features can result in sub-optimal performance (\textit{cf}. Table \ref{tab:3d}); 2) The direct distillation of open-vocabulary features from 2D to 3D can impair the generalizability (\textit{cf}. Table \ref{tab:3d}); 3) The performance of OpenScene-2D significantly deteriorates during inference without depth maps (\textit{cf}. Table \ref{tab:3d}, Figure \ref{fig:qual_3d}).
To address these limitations, we 
leverage Neural Radiance Field (NeRF) to simultaneously encode 3D scene representations and open-vocabulary semantics, where we 
train the model to learn the blending weights of multi-view open-vocabulary features using the supervision from novel views. 
While previous works like LERF \cite{lerf2023}, VL-Fields \cite{vl-fields}, and Open-NeRF \cite{open-nerf} have explored using neural implicit fields to learn open-vocabulary semantics, they still 
require per-scene optimization and cannot generalize to unseen scenes.

In this paper, we propose \textbf{G}eneralizable \textbf{O}pen-\textbf{V}ocabulary \textbf{Ne}ural \textbf{S}emantic \textbf{F}ields (GOV-NeSF), the overall pipeline of which is shown in Figure \ref{teaser}. 
GOV-NeSF is trained using only 2D data without the need for point clouds, ground truth semantic labels or depth maps, and can generalize to unseen scenes for open-vocabulary semantic segmentation. Our model is capable of 2D semantic segmentation from novel views and 3D semantic segmentation of the entire 3D scene. 
Our approach begins with the construction of a cost volume through back-projection of image features, which is then processed by a 3D U-Net \cite{3dunet, 3dresunet} to extract geometry-aware features of the 3D scene. 
Subsequently, during volume rendering, we predict the color and semantic of the sampled 3D points through proposing a Multi-view Joint Fusion Module, which is trained to blend both the color and open-vocabulary values from multi-view projections. 
Additionally, a Cross-View Attention module is introduced to effectively aggregate multi-view image features before the prediction of blending weights. 
Extensive experiments validate our state-of-the-art performance on both 2D and 3D open-vocabulary semantic segmentation. Our \textbf{contributions} can be summarized as:

\begin{enumerate}
    \item To the best of our knowledge, we are the first to explore Generalizable Open-Vocabulary Neural Semantic Fields. 
    Its robust design allows for direct inference in unseen scenes and seamless adaptation across datasets.
    \item The Multi-view Joint Fusion module, a key innovation of our model, blends colors and open-vocabulary features from multi-view inputs. It employs implicit scene representation to predict geometry-aware blending weights and integrates a cross-view attention module for enhanced multi-view feature aggregation.
    \item Extensive experiments demonstrate our state-of-the-art open-vocabulary semantic segmentation results with remarkable generalizability across scenes and datasets.
\end{enumerate}

\section{Related Work}

\textbf{Generalizable NeRF.} 
The field of neural implicit representation has seen significant advances \cite{nerf, neus, instant, mip}, yet these methods depend on computationally intensive per-scene optimization. To overcome this limitation, 
several recent methods have been proposed to place emphasis on the generalization to unseen scenes. 
Particularly, MVSNeRF \cite{mvsnerf}, IBRNet \cite{wang2021ibrnet}, Point-NeRF \cite{xu2022point}, and PixelNeRF \cite{yu2020pixelnerf} are designed to acquire neural radiance fields using images from arbitrary unseen scenes, achieving novel view synthesis without per-scene optimization. Notably, PixelNeRF \cite{yu2020pixelnerf} and IBRNet \cite{wang2021ibrnet} employ volume rendering techniques through back-projecting features from nearby reference images into 3D space. 
Instead of inputting only the nearby views of the source view, we feed our model with the coarsely captured images from the entire unseen scene, in order to simultaneously encode the representation and the open-vocabulary semantics of the whole 3D scene. 

\begin{figure*}[t!]
    \centering
    \includegraphics[width=1\textwidth]{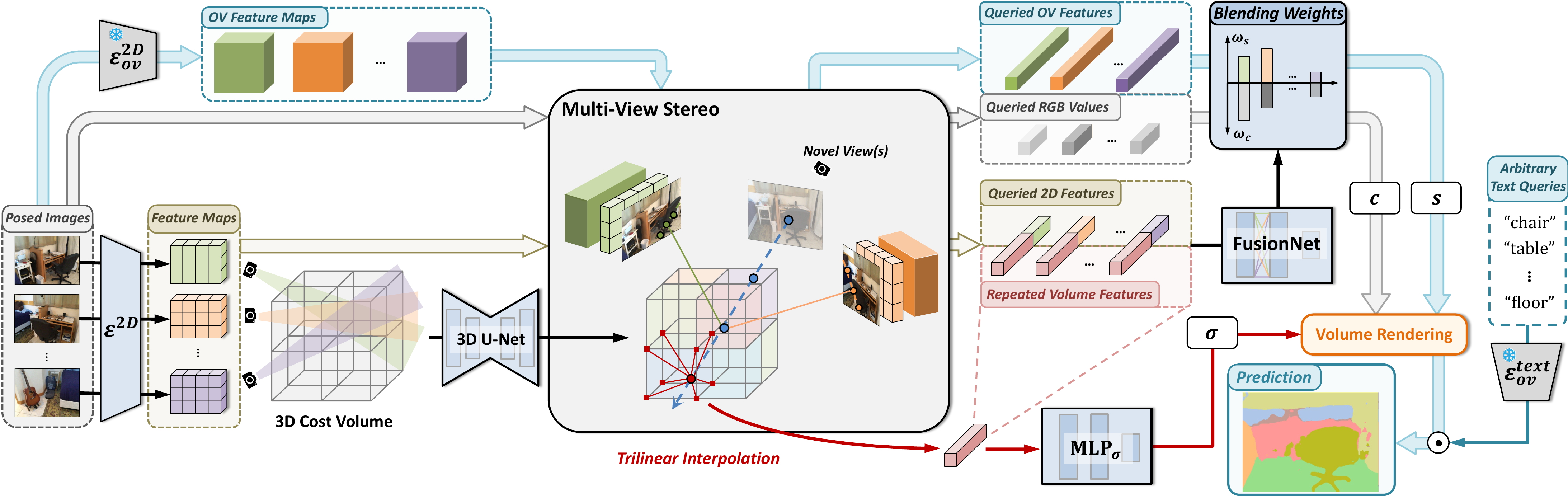}
    \caption{\textbf{Structure of GOV-NeSF.} Given a set of posed images of the 3D scene, we first use a shared image encoder to extract the 2D feature maps, and unproject them to build a 3D cost volume. Moreover, we 
    leverage LSeg \cite{lseg} to predict the per-pixel open-vocabulary features. 
    We then perform Multi-View Stereo to query the 2D 
    and 
    open-vocabulary features for each sampled 3D point along the ray, concatenate the queried 2D features and the volume feature, and feed them into the FusionNet to predict blending weights. The final color and open-vocabulary feature are the weighted sum of multi views using the blending weights.}
    \label{framework}
\end{figure*}

\issue{Neural Semantic Fields.} 
As introduced in \cite{semantic_nerf}, Semantic-NeRF marked a significant milestone in neural implicit representation field by incorporating semantics into the NeRF framework. 
Subsequent researchers have expanded upon this foundational concept. 
For 
example, several studies \cite{fu2022panoptic,kundu2022panoptic,kobayashi2022distilledfeaturefields} extended 
Semantic-NeRF by incorporating instance-level modeling, and \cite{tschernezki22neural} introduced abstract visual features for post hoc semantic segmentation derivation. \cite{yang2021learning} presented the concept of an object-compositional neural radiance field. 
Panoptic NeRF \cite{fu2022panoptic} is designed for panoptic radiance fields to address tasks such as label transfer and scene editing, and GNeSF \cite{chen2023gnesf} introduces a generalizable neural semantic field using a soft voting mechanism. Recently, people also explored distilling the Vision-Language Models knowledge into neural implicit representation to achieve open-vocabulary neural semantic fields. LERF \cite{lerf2023}, VL-Fields \cite{vl-fields} and Open-NeRF \cite{open-nerf} leverage vision-language models to encode the open-vocabulary semantics using neural implicit fields, performing semantic novel view synthesis given arbitrary text query input. Nonetheless, most existing works require per-scene optimization and cannot generalize to unseen scenes.
In contrast to the existing methods, we focus on the development of generalizable open-vocabulary neural semantic fields 
to equip neural semantic fields with the capability to recognize open-world categories across unseen scenes.

\issue{Open-Vocabulary 3D Semantic Segmentation.} With rapid advancements of Vision-Language Models \cite{clip, lseg, ovseg, decoupling, openseg}, several works have proposed to distill the open-vocabulary semantic knowledge into 3D models. One line of works \cite{pointclip, clip2point, ofa, pla} distill image-level CLIP features into the 3D semantic segmentation models, which however 
can suffer from coarse supervision. Recently, OpenScene \cite{openscene} proposes to leverage LSeg \cite{lseg}, OpenSeg \cite{openseg} to fuse the dense pixel-wise open-vocabulary features into the point clouds, and achieves remarkable zero-shot and open-vocabulary 3D semantic segmentation results comparing to existing methods. However, OpenScene requires point clouds input during training and inference, and their fusion of point-wise features is naive averaging.  Moreover, directly distilling the 2D open-vocabulary features into 3D models can lead to limited generalizability, since the 3D model is typically trained on 3D datasets that are remarkably small comparing to 2D web-scale datasets. In contrast, we propose to train a neural semantic fields, in order to learn the multi-view features blending instead of direct distillation, thus further unleashing the generalizability of 2D VLMs.


\section{Methodology}

\subsection{Overview}

The overall framework of our proposed GOV-NeSF is shown in Figure \ref{framework}. Given posed images of a 3D scene, we first use an off-the-shelf 2D Open-Vocabulary semantic segmentation model LSeg \cite{lseg} to extract the per-pixel open-vocabulary feature maps. We also train an image encoder to embed image features, which is followed by back-projection to build a 3D cost volume. 
Subsequently, during volume rendering, we leverage Multi-View Stereo (MVS) to query features for each sampled 3D point, and propose a FusionNet to blend the multi-view colors and open-vocabulary features with cross-view attention.

\subsection{Feature Extraction}

\noindent\textbf{Feature Fusion.} Formally, given a set of posed images $\{\bm{I}^n\}_{n=1}^{N},\ \bm{I}\in\mathbb{R}^{3\times h\times w}$, we train an image encoder $\bm{\epsilon}^{2D}$ to extract the 2D feature maps $\{\bm{F}^n\}_{n=1}^{N},\ \bm{F}^n\in\mathbb{R}^{c\times \bar{h}\times \bar{w}}$, and build a 3D Cost Volume through unprojection. Specifically, for a cost volume $\bm{V}_{c}\in\mathbb{R}^{C\times H\times W\times D}$, each voxel feature is accumulated as:
\begin{equation}
    \bm{v}[:,i,j,k]=[\mathcal{A}_{i,j,k}, \mathcal{V}_{i,j,k}],
\end{equation}

\noindent where $[\ ]$ denotes concatenation, $\mathcal{A},\,\mathcal{V}$ represent average and variance of the features that are unprojected to the voxel. We then use a 3D U-Net \cite{3dunet, 3dresunet} to extract the geometry features of the 3D scene and derive the aggregated Volume Features $\bm{V}_a\in\mathbb{R}^{C'\times H\times W\times D}$.

\issue{Open-Vocabulary Feature Extraction.} To leverage 2D vision-language models to provide dense open-vocabulary features, we use a pre-trained LSeg \cite{lseg} model to predict Open-Vocabulary (OV) feature maps $\{\bm{F}^n_{ov}\}_{n=1}^{N}$, where $\bm{F}^n_{ov}\in\mathbb{R}^{d\times h\times w}$ share the same image height $h$ and width $w$ as $\bm{I}^n$, with per-pixel $d$-dimensional OV feature. 

\begin{figure}[t!]
    \centering
    \includegraphics[width=1\linewidth]{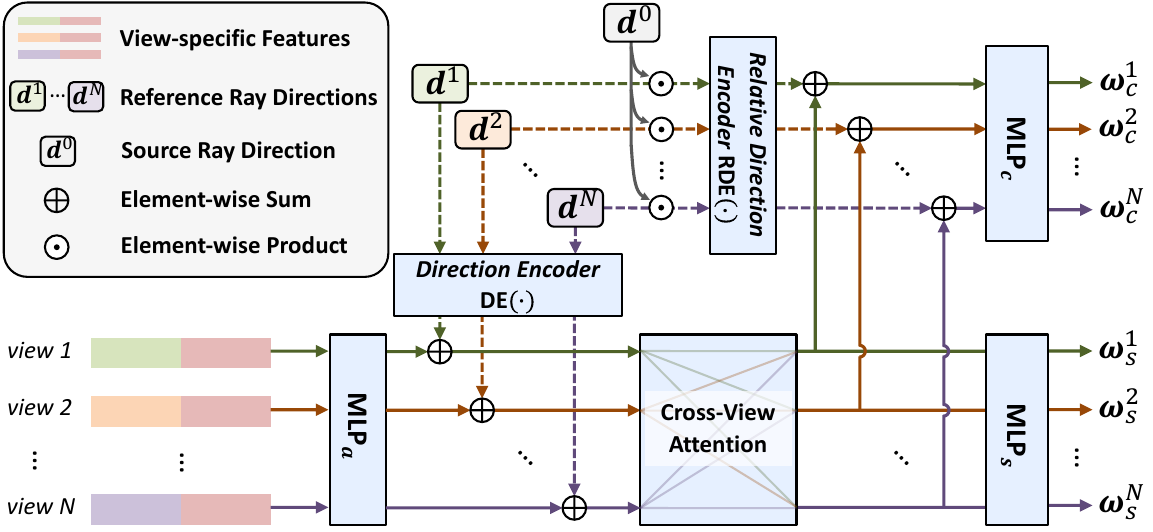}
    \caption{\textbf{FusionNet Structure.} We aggregate multi-view features through Cross-View Attention module, and predict view-specific blending weights. Refer to the text for more details.}
    \label{fusionnet}
\end{figure}

\subsection{Multi-view Joint Fusion}

\noindent\textbf{Multi-View Stereo.} As shown in Figure \ref{fusionnet}, given the input posed images $\{\bm{I}^n\}_{n=1}^{N}$ as reference images,
and their 2D feature maps $\{\bm{F}^n\}_{n=1}^{N}$ and OV feature maps $\{\bm{F}^n_{ov}\}_{n=1}^{N}$, we perform volume rendering to render images from novel views. 
Given a ray $\{\bm{r}(t)=\bm{o}+t\bm{d}, t\geq 0\}$ from the source view, we project a 3D point $\bm{p}$ sampled on the ray onto the reference views and get its normalized 2D coordinates $\{\pi_n(\bm{p})\}_{n=1}^{N}$.
We then
use bilinear interpolation to get the queried RGB values $\{\bm{i}^n=\bm{I}^n(\pi_n(\bm{p}))\}_{n=1}^{N}$, 2D features $\{\bm{f}^n=\bm{F}^n(\pi_n(\bm{p}))\}_{n=1}^{N}$, and OV feature $\{\bm{f}_{ov}^n=\bm{F}_{ov}^n(\pi_n(\bm{p}))\}_{n=1}^{N}$. Furthermore, we 
query the volume feature at $\bm{p}$ using trilinear interpolation to get $\bm{v}=\bm{V}_a(:, \bm{p})$.

\issue{Joint View Blending.} Since we are targeting generalizable neural implicit fields for room-scale scene representation, it is difficult to directly regress 
colors and 
OV features in the 3D space (
\cf Table \ref{tab:2d}, Table \ref{tab:3d}, Table \ref{tab:nvs}). 
We thus propose to jointly blend the colors and the OV features from reference views instead of direct regression. Specifically, we construct the view-specific features for $n$-th view as $\bm{x}^n = [\bm{f}^n,\ \bm{v}]$, and feed $\{\bm{x}^n\}_{n=1}^{N}$ into our proposed FusionNet to predict view-specific weights:
\begin{equation}
    \{\bm{w}_c^n\}_{n=1}^{N},\ \{\bm{w}_s^n\}_{n=1}^{N} = \text{FusionNet}(\{\bm{x}^n\}_{n=1}^{N}),
\end{equation}
where $\{\bm{w}_c^n\}_{n=1}^{N}$, $\{\bm{w}_s^n\}_{n=1}^{N}$ are the blending weights for colors and OV features, respectively. We then normalize them using Softmax to get $\{\bar{\bm{w}}_c^n\}_{n=1}^{N}$ and $\{\bar{\bm{w}}_s^n\}_{n=1}^{N}$, and compute the color $\bm{c}$ and OV feature $\bm{s}$ of 3D point $\bm{p}$ as:
\begin{equation}
    \bm{c}=\sum_{n=1}^{N}\bar{\bm{w}}_c^n\cdot\bm{i}^n,\;\;\;\;\bm{s}=\sum_{n=1}^{N}\bar{\bm{w}}_s^n\cdot\bm{f}^n_{ov}.
\end{equation}
%
Additionally, we predict the density of $\bm{p}$ from the volume feature: $\sigma=\text{MLP}_{\sigma}(\bm{v})$.

\issue{FusionNet.} As shown in Figure \ref{fusionnet}, given the cross-view features $\{\bm{x}^n\}_{n=1}^{N}$, we first use a shared tiny MLP to aggregate features for each view: $\{\tilde{\bm{x}}^n\}_{n=1}^{N} = \text{MLP}_{a}(\{\bm{x}^n\}_{n=1}^{N})$. 
Subsequently, we propose a Cross-View Attention (CVA) module based on self-attention \cite{attention} 
to exchange information across views before blending weights prediction:
\begin{equation}
    \{\tilde{\bm{x}}^n_{att}\}_{n=1}^{N} = \text{CVA}\left(\left\{\tilde{\bm{x}}^n+\text{DE}\left(\bm{d}^n\right)\right\}_{n=1}^{N}\right),
\end{equation}
where CVA($\cdot$) is a self-attention module based on Transformers \cite{attention, vit}, $\bm{d}^n$ is the ray direction from $n$-th camera to $\bm{p}$, and $\text{DE}(\cdot)$ is the ray direction encoder. 

To equip the color prediction with source-view-dependency, we compute the color blending weights using:
\begin{equation}
    \{\bm{w}_c^n\}_{n=1}^{N}=\text{MLP}_{c}\left(\left\{\tilde{\bm{x}}^n_{att}+\text{RDE}\left(\bm{d}^0\cdot \bm{d}^n\right)\right\}_{n=1}^{N}\right),
\end{equation}
where $\bm{d}^0$ is the ray direction from source camera to $\bm{p}$, and $\text{RDE}(\cdot)$ is relative ray direction encoder. 
Furthermore, the prediction of OV feature blending weights should be source-view-agnostic:
\begin{equation}
    \{\bm{w}_s^n\}_{n=1}^{N}=\text{MLP}_{s}\left(\left\{\tilde{\bm{x}}^n_{att}\right\}_{n=1}^{N}\right).
\end{equation}

\issue{Joint Volume Rendering.} Similarly to volume rendering \cite{nerf}, we accumulate both the colors and OV features along each ray using the shared density field:
\begin{subequations}
    \begin{equation}
        \hat{\bm{C}}(\bm{r})=\sum_{i=1}^{N_p}T_i\left(1-\text{exp}(-\sigma_i\delta_i)\right)\bm{c}_i,
    \end{equation}
    \begin{equation}
        \hat{\bm{S}}(\bm{r})=\sum_{i=1}^{N_p}T_i\left(1-\text{exp}(-\sigma_i\delta_i)\right)\bm{s}_i,
    \end{equation}
    \begin{equation}
        T_i=\text{exp}\left(-\sum_{j=1}^{i-1}\sigma_j\delta_j\right),
    \end{equation}
\end{subequations}
where $\delta_j$ is the distance between sampled points along the ray. $\{\bm{c}_i\}_{i=1}^{N_p}$, $\{\bm{s}_i\}_{i=1}^{N_p}$, $\{\sigma_i\}_{i=1}^{N_p}$ are the colors, OV features, and densities of the points along the ray.


\subsection{Training Objective}

Our loss function is defined as:
\begin{equation}
    \mathcal{L}=\mathcal{L}_{color} + \alpha\mathcal{L}_{ov},
\end{equation}
where $\alpha$ is the weight balancing two loss terms. The color loss $\mathcal{L}_{color}$ and open-vocabulary feature loss $\mathcal{L}_{ov}$ are respectively defined as:
\begin{subequations}
    \begin{equation}
        \mathcal{L}_{color}=\frac{1}{|\mathcal{R}|}\sum_{\bm{r}\in\mathcal{R}}\left\lVert\hat{\bm{C}}(\bm{r})-\bm{C}(\bm{r})\right\rVert_2^2,
    \end{equation}
    \begin{equation}
        \mathcal{L}_{ov}=\frac{1}{|\mathcal{R}|}\sum_{\bm{r}\in\mathcal{R}}\left(1-\text{cos}\left(\hat{\bm{S}}(\bm{r}), \bm{S}(\bm{r})\right)\right),
    \end{equation}
\end{subequations}
\noindent where $\bm{C}(\bm{r})$ and $\bm{S}(\bm{r})$ are the ground truth color and OV feature for ray $\bm{r}$, and $\text{cos}(\cdot)$ is cosine similarity. Note that since the OV feature map prediction from LSeg is not view-consistent, we detach the gradient from $\mathcal{L}_{ov}$ to $\bm{\sigma}$ to enhance the quality of the learned density fields.

\subsection{Inference}

Given input posed images of a 3D scene that is unseen during training, we first encode them into our framework to build the scene representation with OV semantics. 
We then input an arbitrary set of texts $\{\bm{t}^k\}_{k=1}^{K}$ and encode the text features using the frozen text encoder $\bm{\epsilon}_{ov}^{text}$: $\{\bm{\epsilon}_{ov}^{text}(\bm{t}^k)\}_{k=1}^{K}$, 
which are utilized for both 2D and 3D Open-Vocabulary Semantic Segmentation.

\issue{2D semantic segmentation.} We 
render the OV feature maps from arbitrary novel views: $\{\hat{\bm{F}}_{ov}^m\}_{m=1}^{M}$ and compute its pixel-wise cosine similarity with $\{\bm{f}_t(\bm{t}^k)\}_{k=1}^{K}$. 
The per-pixel segmentation result is then given by the $\text{argmax}$ within the similarities with all the query text features.

\issue{3D semantic segmentation.} Our model is also capable of segmenting any given point clouds corresponding to the input images. We simply query the source-view-agnostic OV features of the given point coordinates and perform per-point segmentation similarly as 2D semantic segmentation. Note that comparing to OpenScene-2D \cite{openscene} which uses naive averaging between multi-view OV feature maps, our model leverages neural implicit representation to learn the blending weights based on the supervision from novel views OV feature maps.

\issue{Remarks.} Although our method does not require the ground truth depth maps during training, we can also leverage the depth maps during inference similarly as OpenScene-2D \cite{openscene} through a Depth Guided Masking (DGM) method. We compare the distances from 3D points to multi views, and mask out the blending weights of the views where the distances differ over $25\%$ from ground truth depths. In the subsequent experiments, ``Ours" does not include the DGM module unless otherwise specified.

\section{Experiments}

We conduct extensive experiments on 2D and 3D generalizable open-vocabulary semantic segmentation tasks. 
We also compare our novel view synthesis performance with existing generalizable NeRF methods to quantify our scene representation quality. Furthermore, we perform detailed qualitative comparisions on both 2D and 3D open-vocabulary semantic segmentation, and conduct extensive ablation studies on our proposed components.

\subsection{Experiment Setup}

\noindent\textbf{Implementation Details. } For 2D feature extraction, we leverage ResUNet-34 \cite{resunet} to encode 2D feature maps, and use the ViT-L/16-based LSeg \cite{lseg} model to extract OV feature maps. For 3D feature aggregation, we leverage a 3D ResUNet \cite{3dresunet} containing 3 levels. For volume rendering, we sample 64 points along each ray and do not perform hierarchical sampling to reduce GPU consumption. 
\begin{table*}[htbp]
    \small
    \centering
    \setlength{\tabcolsep}{4mm}{
    \begin{tabular}{c|c|ccc|ccc}
        \toprule

         &\multirow{2}{*}{Method}&\multicolumn{3}{c|}{ScanNet \cite{dai2017scannet}} & \multicolumn{3}{c}{Replica \cite{replica}}\\
         & & mIoU& oAcc& mAcc& mIoU& oAcc& mAcc\\
         \midrule[0.8pt]
         \midrule
         \multirow{3}{*}{\textit{Fully-Supervised}} & MVSNeRF \cite{mvsnerf}+Semantic Head$^\dag$& 39.8 & 60.0 & 46.0 & 23.4 & 54.3 & 33.7\\
         &NeuRay \cite{neuray}+Semantic Head$^\dag$& 51.0  & 77.6 & 57.1  & 35.9 & 69.4 & 44.0\\
         & S-Ray \cite{semanticray}$^\dag$& 57.2 & 78.2 & 62.6 & 41.6 & 70.5 & 47.2 \\
         \midrule
         \multirow{5}{*}{\textit{Open-Vocabulary}} &S-Ray \cite{semanticray}-OV& 33.9 & 50.6 & 45.7 & 9.7 & 26.7 & 18.2 \\
         &Distill Baseline& 46.4  &  69.0 & 56.5 & 5.0 & 28.1 & 11.5 \\
         &Ours & \textbf{52.2} & \textbf{73.8} & \textbf{62.2} & \textbf{44.3} & \textbf{76.2} & \textbf{57.6}  \\
         \cmidrule{2-8}
         &LSeg$_{rd}$ \cite{lseg}$^\ddag$ & 48.4 & 69.4 & 57.6 & 23.2 & 53.7 & 31.3 \\
         &LSeg$_{gt}$ \cite{lseg}$^\ddag$ & 55.9 & 77.3 & 65.4 & 52.0 & 79.6 & 64.9 \\
        
         \bottomrule
    \end{tabular}
    }
    \captionsetup{font=small}
    \caption{\small \textbf{Generalizable NeRF-based 2D Semantic Segmentation Results.} We report semantic segmentation results from novel views in novel scenes. $^\dag$ are results reported in \cite{semanticray}. $^\ddag$ LSeg$_{rd}$, LSeg$_{gt}$ are the results of directly applying LSeg \cite{lseg} on the rendered images by NeuRay \cite{neuray} or ground truth images at novel views, respectively.}
    \label{tab:2d}
\end{table*}
\begin{table*}[htbp]
    \small
    \centering
    \setlength{\tabcolsep}{2.6mm}{
    \begin{tabular}{c|c|ccc|ccc}
        \toprule

         \multirow{2}{*}{Model Input}&\multirow{2}{*}{Method}&\multicolumn{3}{c|}{ScanNet \cite{dai2017scannet}} & \multicolumn{3}{c}{Replica \cite{replica}}\\
         & & 3D mIoU&3D oAcc& 3D mAcc& 3D mIoU&3D oAcc&3D mAcc\\
         \midrule[0.8pt]
         \midrule
         3D & OpenScene-3D \cite{openscene}$^\dag$ & 51.6 & 72.8 & 62.5 & 3.3& 27.7 & 6.5 \\
         \midrule
         \multirow{3}{*}{2D \textit{w/o Depth}}& OpenScene-2D \cite{openscene} \textit{w/o} Depth$^\dag$& 42.1 &64.4 &56.8  & 31.2 & 64.4 & 48.7  \\
         &Distill Baseline& 43.4 & 64.0& 56.2 & 17.7 & 42.7 & 31.0 \\
         &Ours & \textbf{45.7}& \textbf{68.3} & \textbf{60.3} & \textbf{32.8} & \textbf{66.3} & \textbf{49.4}\\
         \midrule
         \multirow{2}{*}{2D \textit{w/ Depth}} &OpenScene-2D \cite{openscene}$^\dag$& 52.2& 73.4 & 63.3 & 35.8 &  69.4 & 52.2\\
         &Ours \textit{w/} DGM & \textbf{53.5}& \textbf{74.6}& \textbf{64.4} & \textbf{37.7} & \textbf{71.5} & \textbf{53.3}\\
        
         \bottomrule
    \end{tabular}
    }
    \captionsetup{font=small}
    \caption{\small \textbf{Open-Vocabulary 3D Semantic Segmentation Results.} $^\dag$ are our reproduced results. 
    }
    \label{tab:3d}
\end{table*}
\begin{table*}[h!]
    \small
    \centering
    \setlength{\tabcolsep}{6.67mm}{
    \begin{tabular}{c|ccc|ccc}
        \toprule

         \multirow{2}{*}{Method}&\multicolumn{3}{c|}{ScanNet \cite{dai2017scannet}} & \multicolumn{3}{c}{Replica \cite{replica}}\\
         & PSNR$\uparrow$&SSIM$\uparrow$& LPIPS$\downarrow$& PSNR$\uparrow$&SSIM$\uparrow$& LPIPS$\downarrow$\\
         \midrule[0.8pt]
         \midrule
         NeRF-Det \cite{nerfdet}$^\dag$& 18.8  & 0.743 & 0.505 & 13.5 & 0.664 & 0.618 \\
         NeuRay \cite{neuray}$^\dag$ & \textbf{22.8} & 0.786 & \textbf{0.142} & 18.1 & 0.606 & \textbf{0.195} \\
         S-Ray \cite{semanticray}$^\dag$&  22.0 & 0.769 & 0.154  & 18.0 &  0.604 & 0.203  \\
         Ours & 21.5 & \textbf{0.801} & 0.456 & \textbf{21.5}& \textbf{0.811} & 0.419\\
        
         \bottomrule
    \end{tabular}
    }
    \captionsetup{font=small}
    \caption{\small \textbf{Novel View Synthesis Results.} $^\dag$ are our reproduced results using our extracted datasets.}
    \label{tab:nvs}
\end{table*}

\issue{Datasets.} 
We mainly conduct experiments on two datasets: real-world dataset ScanNet \cite{dai2017scannet}, and synthetic dataset Replica \cite{replica}. ScanNet is a large RGB-D dataset containing 2.5M views in 1,513 real-world 3D indoor scenes with the corresponding camera poses and semantic labels. 
We train our models on their provided training set split and evaluate on the validation set. 
Replica provides 18 high-quality synthetic 3D indoor scenes with mesh and ground truth 2D semantic labels. We follow Semantic-NeRF \cite{semantic_nerf} to generate sets of posed images in 8 scenes. Since Replica \cite{replica} does not provide the ground truth 3D semantic labels, we leverage the TSDF Fusion in \cite{sun2021neucon} to generate the mesh with semantic labels. To evaluate the generalizability of our proposed model, we train our models on ScanNet train set and evaluate them on \textit{both} ScanNet val set and Replica. For each scene, we extract 100 images with corresponding camera poses. During training, 
we input 30 posed images as reference images and render 3 images from novel views for each iteration. During testing, 
we input 95 images and render 5 images from novel views for evaluating 2D semantic segmentation, and input all 100 images for evaluating 3D semantic segmentation.

\issue{Metrics.} To evaluate the semantic segmentation quality, we compute 
mean Intersection-over-Union (mIoU), total Accuracy (oAcc), and average Accuracy (mAcc) on both 2D and 3D semantic segmentation. 
Similar to previous works 
\cite{nerf, wang2021ibrnet},  we report 
PSNR, SSIM \cite{ssim} and LPIPS$_{vgg}$ \cite{lpips} to evaluate the quality of our novel view synthesis.

\issue{Baselines.} 
We build the 2D and 3D baselines since there is no prior work 
on generalizable open-vocabulary neural semantic fields:
\begin{itemize}
    \item \textbf{2D Semantic Segmentation.} We modify the model of S-Ray \cite{semanticray} to S-Ray-OV, which predicts open-vocabulary feature maps instead of one-hot semantic logits to reproduce their open-vocabulary results. We also compare with the Distill Baseline, which shares the basic framework design as ours without the Multi-view Joint Fusion module, \textit{i.e.} directly regressing open-vocabulary features using the multi-view image features and volume features.
    \item \textbf{3D Semantic Segmentation.} We mainly compare with the Distill Baseline, and LSeg \cite{lseg}-based OpenScene-2D and OpenScene-3D \cite{openscene}. For a comprehensive analysis, we compare with OpenScene-2D in both \textit{w/} and \textit{w/o} Depth scenarios. For evaluating OpenScene-3D on Replica, we use their provided model pre-trained on ScanNet with fused LSeg features. We first input their preprocessed point clouds of Repilca scenes to predict the dense OV features, and transfer them to the meshes that we extracted using TSDF Fusion.
\end{itemize}
Furthermore, we still compare with NeRF-Det \cite{nerfdet}, NeuRay \cite{neuray} and S-Ray \cite{semanticray} in terms of novel view synthesis to evaluate the quality of our 3D scene representations despite novel view synthesis is not the focus of GOV-NeSF.

\begin{figure*}[t!]
    \centering
    \includegraphics[width=1\textwidth]{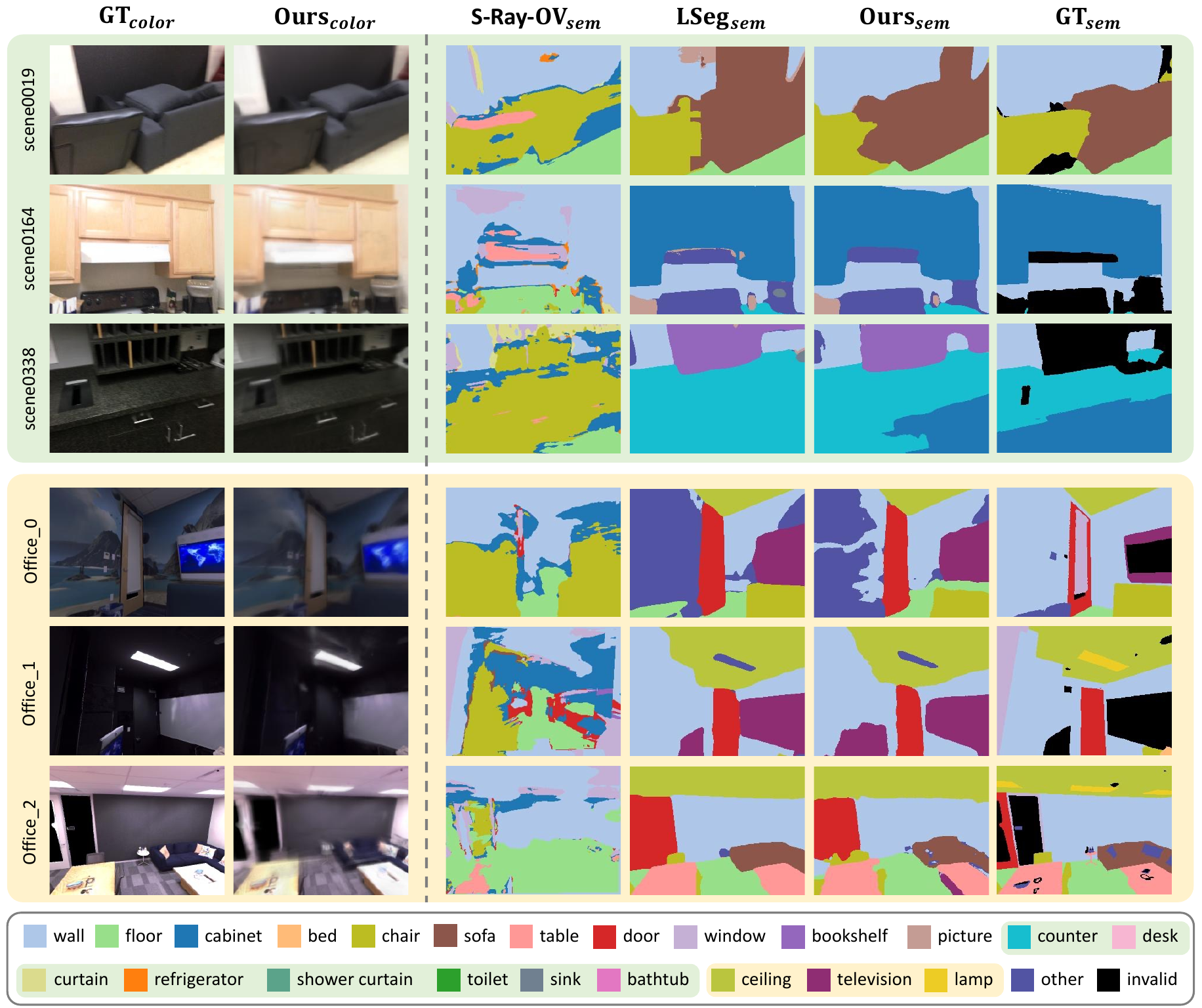}
    \caption{\textbf{Visualization of 2D results.} We show the GT color images, our rendered color images, S-Ray \cite{semanticray}-OV rendered semantics, our rendered semantics, LSeg$_{gt}$ \cite{lseg} predictions, and GT semantics on novel views from unseen scenes in \colorbox{scannet}{ScanNet} \cite{dai2017scannet} and \colorbox{replica}{Replica} \cite{replica}.} 
    \label{fig:qual_2d}
\end{figure*}
\subsection{Main Results}
\noindent\textbf{2D Semantic Segmentation.} In Table \ref{tab:2d}, we show semantic segmentation results of various Generalizable NeRF-based methods. Our Distill Baseline surpasses S-Ray \cite{semanticray}-OV and performs comparably to LSeg$_{rd}$ \cite{lseg} on ScanNet. 
However, when applied to the Replica dataset, the Distill Baseline shows significantly reduced effectiveness, highlighting the undermined generalizability in direct distillation. 
In contrast, our method exhibits a substantial improvement on ScanNet, outperforming S-Ray-OV by a margin of $+18.3$ in mIoU and the Distill Baseline by $+5.8$ mIoU. More notably, on the Replica dataset, our approach exceeds both the S-Ray-OV and Distill Baseline by over $+34.6$ mIoU and $+39.4$ in mAcc. These significant improvements can be attributed to our Joint Blending module, which blends multi-view OV features instead of direct regression. 
Furthermore, our approach outperforms LSeg$_{rd}$ \cite{lseg} by $+3.8$ mIoU in ScanNet and $+21.1$ mIoU in Replica. 
Note that our given images are relatively sparse, which results in challenges in rendering high-quality images from novel views, yet our performance remains comparble to LSeg$_{gt}$ when transferred to Replica without fine-tuning.

\issue{3D Semantic Segmentation.} Table \ref{tab:3d} shows the comparisons with different open-vocabulary 3D semantic segmentation methods. OpenScene-3D performs comparably to OpenScene-2D and our method in ScanNet, yet greatly degenerates when transferred to Replica, necessitating the usage of 2D VLMs during inference. Under the \textit{w/o Depth} setting, we reproduce OpenScene-2D \cite{openscene} results without 
depth-based masking. Our Distill Baseline performs comparably as the OpenScene-2D \textit{w/o Depth} on ScanNet, and our approach surpasses both baselines by at least $+2.3$ mIoU on ScanNet and $+1.6$ mIoU on Replica. Notably, our approach surpasses OpenScene-3D and Distill Baseline by $+29.5$ mIoU and $+15.1$ mIoU on Replica, respectively, thus further emphasizing the robustness of our model. 
Additionally, our method can also leverage depth maps during inference when available to further improve OpenScene-2D performance by $+1.3\sim +1.9$ mIoU. 
\begin{figure*}[t!]
    \centering
    \includegraphics[width=1\textwidth]{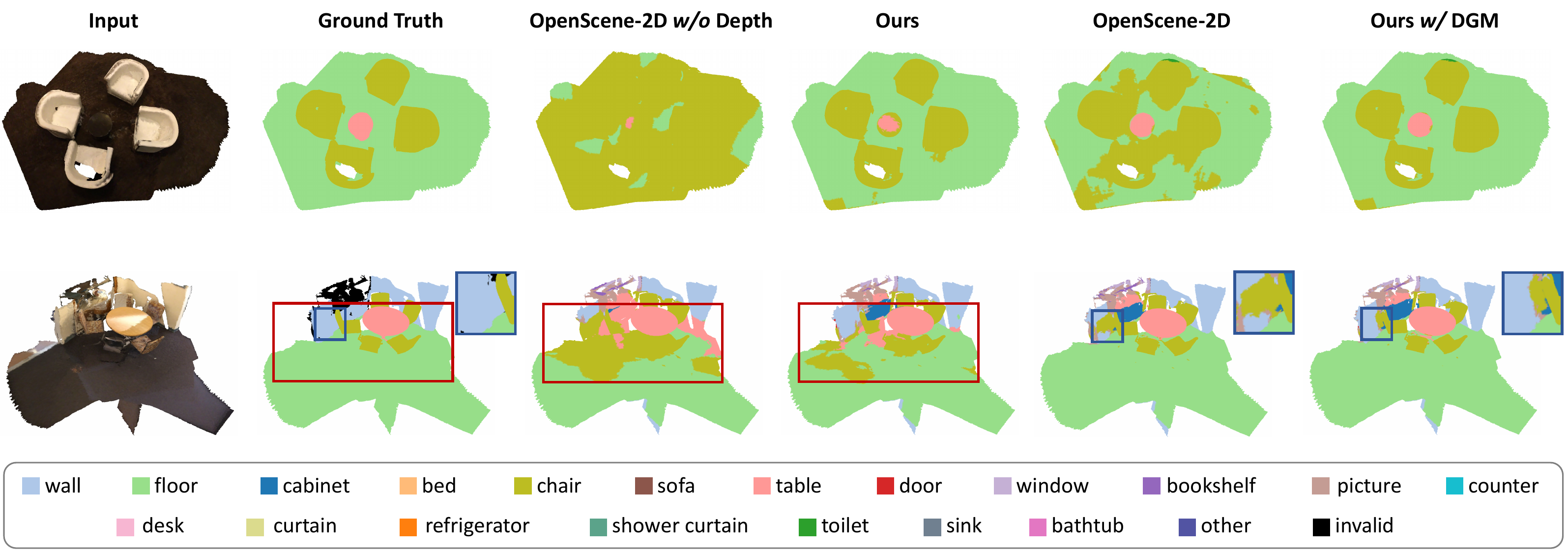}
    \caption{\textbf{Visualization of 3D results.} We compare with OpenScene-2D \cite{openscene} in terms of 3D semantic segmentation on ScanNet.} 
    \label{fig:qual_3d}
\end{figure*}

\begin{table*}[h!]
    \small
    \centering
        \setlength{\tabcolsep}{5.1mm}
        {
        \begin{tabular}{ccc|c|ccc|ccc}
        \toprule
         \multirow{2}{*}{MJF}& \multirow{2}{*}{VFA} & \multirow{2}{*}{CVA} & \multirow{2}{*}{DGM} & 
         \multicolumn{3}{c|}{2D \textit{Seg.}}&\multicolumn{3}{c}{3D \textit{Seg.}}\\
         \cmidrule{5-10}
         & &&&mIoU &oAcc& mAcc& mIoU &oAcc& mAcc \\
         \midrule
         &\checkmark&&&   46.4 & 69.0 & 56.5 & 39.4 & 62.2& 54.1 \\
         \checkmark&&& & 49.5 & 71.8 & 60.4 & 44.2 & 67.1 & 57.5  \\
         \checkmark&\checkmark&& & 50.7 & 72.1 & 60.9 & 44.6 & 67.4 & 57.8 \\
         \checkmark&\checkmark&\checkmark&& \textbf{52.2}& \textbf{73.8} & \textbf{62.2}&45.7 & 68.3 & 59.1  \\
         \checkmark&\checkmark&\checkmark&\checkmark& 46.8 & 69.8& 58.3 &\textbf{53.5} & \textbf{74.6}  &\textbf{64.2} \\

         \bottomrule
         
        \end{tabular}}
        \captionsetup{font=small}
        \caption{\small \textbf{
        Ablation study} on ScanNet, evaluating contributions of our proposed components.
        } 
        \vspace{-1mm}
        \label{tab:ablate}
\end{table*}
\issue{Novel View Synthesis.} We also evaluate the novel view synthesis quality in Table \ref{tab:nvs}, where our approach achieves superior performance than NeRF-Det \cite{nerfdet} and comparable results as NeuRay \cite{neuray} and S-Ray \cite{semanticray}. Note that both NeuRay and S-Ray leverage the ground truth depth maps during training, while we only learn the density field based on RGB images. 
Moreover, we consistently maintains the quality of scene representation when transferred to Replica.

\subsection{Qualitative Results}

\noindent\textbf{2D results.} We visualize the 2D results in Figure \ref{fig:qual_2d}, where 
color images and open-vocabulary semantics are rendered from novel views in unseen scenes. Our color image renderings are relatively blurry compared to the rendering in S-Ray \cite{semanticray} since we are targeting room-scale representation without depth priors. We demonstrate significant improvements over S-Ray \cite{semanticray}-OV since they fail to effectively regress the OV features through direct distillation. Our results are also comparable to the LSeg$_{gt}$ \cite{lseg}, and in some cases (row 1,3,4) we can surpass their results through the aggregation of multi-view OV features.

\issue{3D results.} We visualize the 3D results in Figure \ref{fig:qual_3d}. Our approach demonstrates significant improvements over OpenScene-2D \cite{openscene} when not given ground truth depth maps, and further refines multi-view OV features fusion when given ground truth depth maps. Our MJF module automatically learns to reason about the occlusion without depth supervision, and facilitates more efficient inference compared to the volume rendering-based depth estimation.

\subsection{Ablation Study}

Table \ref{tab:ablate} shows the ablation study we conducted on ScanNet. 
The first row with the same design as Distill Baseline, greatly suffers from domain gaps and significantly degenerates when transferred to Replica (\textit{cf}. Table \ref{tab:2d}). This demonstrates the necessity of our MJF module to learn the blending weights for OV features. Furthermore, the integration of Volume Feature Aggregation and Cross-View Attention module improves both 2D and 3D semantic segmentation performance. 
This improvement underscores the benefits of aggregating scene-level geometry features and cross-view image features prior to blending weights prediction.
Moreover, the Depth-Guided Masking module can lead to significant improvements on 3D semantic segmentation by explicitly masking out the occluded projections. 
However, it also 
causes a drop in 2D semantic segmentation performance 
since it 
can result in empty holes in the rendered images. 
Consequently, DGM module is only used in 3D semantic segmentation when depth masks are available.


\section{Conclusion}
In this paper, we introduce GOV-NeSF, a pioneering framework for generalizable open-vocabulary neural semantic fields. Leveraging the neural implicit representation, GOV-NeSF is designed to learn the joint blending weights for both colors and open-vocabulary features queried from multi-view images, eliminating the need for 3D data, depth priors, or explicit ground truth semantic labels. 
Our framework design enables GOV-NeSF to excel in open-vocabulary semantic segmentation across both 2D semantic NVS and 3D, and set state-of-the-art in benchmarks.

\section{Acknowledgement}
This work is supported by the Agency for Science, Technology and Research (A*STAR) under its MTC Programmatic Funds (Grant No. M23L7b0021).

{
    \small
    \bibliographystyle{ieeenat_fullname}
    \bibliography{main}

\begin{thebibliography}{57}
\providecommand{\natexlab}[1]{#1}
\providecommand{\url}[1]{\texttt{#1}}
\expandafter\ifx\csname urlstyle\endcsname\relax
  \providecommand{\doi}[1]{doi: #1}\else
  \providecommand{\doi}{doi: \begingroup \urlstyle{rm}\Url}\fi

\bibitem[Barron et~al.(2022)Barron, Mildenhall, Verbin, Srinivasan, and Hedman]{mip}
Jonathan~T Barron, Ben Mildenhall, Dor Verbin, Pratul~P Srinivasan, and Peter Hedman.
\newblock Mip-nerf 360: Unbounded anti-aliased neural radiance fields.
\newblock In \emph{Proceedings of the IEEE/CVF Conference on Computer Vision and Pattern Recognition}, pages 5470--5479, 2022.

\bibitem[Chen et~al.(2021)Chen, Xu, Zhao, Zhang, Xiang, Yu, and Su]{mvsnerf}
Anpei Chen, Zexiang Xu, Fuqiang Zhao, Xiaoshuai Zhang, Fanbo Xiang, Jingyi Yu, and Hao Su.
\newblock Mvsnerf: Fast generalizable radiance field reconstruction from multi-view stereo.
\newblock In \emph{Proceedings of the IEEE/CVF International Conference on Computer Vision}, pages 14124--14133, 2021.

\bibitem[Chen et~al.(2023)Chen, Li, Guo, Yan, and Lee]{chen2023gnesf}
Hanlin Chen, Chen Li, Mengqi Guo, Zhiwen Yan, and Gim~Hee Lee.
\newblock Gnesf: Generalizable neural semantic fields.
\newblock \emph{arXiv preprint arXiv:2310.15712}, 2023.

\bibitem[Choy et~al.(2019)Choy, Gwak, and Savarese]{minkowski}
Christopher Choy, JunYoung Gwak, and Silvio Savarese.
\newblock 4d spatio-temporal convnets: Minkowski convolutional neural networks.
\newblock In \emph{Proceedings of the IEEE/CVF conference on computer vision and pattern recognition}, pages 3075--3084, 2019.

\bibitem[{\c{C}}i{\c{c}}ek et~al.(2016){\c{C}}i{\c{c}}ek, Abdulkadir, Lienkamp, Brox, and Ronneberger]{3dunet}
{\"O}zg{\"u}n {\c{C}}i{\c{c}}ek, Ahmed Abdulkadir, Soeren~S Lienkamp, Thomas Brox, and Olaf Ronneberger.
\newblock 3d u-net: learning dense volumetric segmentation from sparse annotation.
\newblock In \emph{Medical Image Computing and Computer-Assisted Intervention--MICCAI 2016: 19th International Conference, Athens, Greece, October 17-21, 2016, Proceedings, Part II 19}, pages 424--432. Springer, 2016.

\bibitem[Dai et~al.(2017)Dai, Chang, Savva, Halber, Funkhouser, and Nie{\ss}ner]{dai2017scannet}
Angela Dai, Angel~X Chang, Manolis Savva, Maciej Halber, Thomas Funkhouser, and Matthias Nie{\ss}ner.
\newblock Scannet: Richly-annotated 3d reconstructions of indoor scenes.
\newblock In \emph{Proceedings of the IEEE conference on computer vision and pattern recognition}, pages 5828--5839, 2017.

\bibitem[Diakogiannis et~al.(2020)Diakogiannis, Waldner, Caccetta, and Wu]{resunet}
Foivos~I Diakogiannis, Fran{\c{c}}ois Waldner, Peter Caccetta, and Chen Wu.
\newblock Resunet-a: A deep learning framework for semantic segmentation of remotely sensed data.
\newblock \emph{ISPRS Journal of Photogrammetry and Remote Sensing}, 162:\penalty0 94--114, 2020.

\bibitem[Ding et~al.(2021)Ding, Xue, Xia, and Dai]{decoupling}
Jian Ding, Nan Xue, Guisong Xia, and Dengxin Dai.
\newblock Decoupling zero-shot semantic segmentation. 2022 ieee.
\newblock In \emph{CVF Conference on Computer Vision and Pattern Recognition (CVPR)}, pages 11573--11582, 2021.

\bibitem[Ding et~al.(2023)Ding, Yang, Xue, Zhang, Bai, and Qi]{pla}
Runyu Ding, Jihan Yang, Chuhui Xue, Wenqing Zhang, Song Bai, and Xiaojuan Qi.
\newblock Pla: Language-driven open-vocabulary 3d scene understanding.
\newblock In \emph{Proceedings of the IEEE/CVF Conference on Computer Vision and Pattern Recognition}, pages 7010--7019, 2023.

\bibitem[Dosovitskiy et~al.(2010)Dosovitskiy, Beyer, Kolesnikov, Weissenborn, Zhai, Unterthiner, Dehghani, Minderer, Heigold, Gelly, et~al.]{vit}
Alexey Dosovitskiy, Lucas Beyer, Alexander Kolesnikov, Dirk Weissenborn, Xiaohua Zhai, Thomas Unterthiner, Mostafa Dehghani, Matthias Minderer, Georg Heigold, Sylvain Gelly, et~al.
\newblock An image is worth 16x16 words: Transformers for image recognition at scale. arxiv 2020.
\newblock \emph{arXiv preprint arXiv:2010.11929}, 2010.

\bibitem[Fu et~al.(2022)Fu, Zhang, Chen, Lu, Zhu, Zhou, Geiger, and Liao]{fu2022panoptic}
Xiao Fu, Shangzhan Zhang, Tianrun Chen, Yichong Lu, Lanyun Zhu, Xiaowei Zhou, Andreas Geiger, and Yiyi Liao.
\newblock Panoptic nerf: 3d-to-2d label transfer for panoptic urban scene segmentation.
\newblock In \emph{International Conference on 3D Vision (3DV)}, 2022.

\bibitem[Ghiasi et~al.(2022)Ghiasi, Gu, Cui, and Lin]{openseg}
Golnaz Ghiasi, Xiuye Gu, Yin Cui, and Tsung-Yi Lin.
\newblock Scaling open-vocabulary image segmentation with image-level labels.
\newblock In \emph{European Conference on Computer Vision}, pages 540--557. Springer, 2022.

\bibitem[He et~al.(2016)He, Zhang, Ren, and Sun]{2d2}
Kaiming He, Xiangyu Zhang, Shaoqing Ren, and Jian Sun.
\newblock Deep residual learning for image recognition.
\newblock In \emph{Proceedings of the IEEE conference on computer vision and pattern recognition}, pages 770--778, 2016.

\bibitem[Huang et~al.(2023)Huang, Dong, Yang, Huang, Lau, Ouyang, and Zuo]{clip2point}
Tianyu Huang, Bowen Dong, Yunhan Yang, Xiaoshui Huang, Rynson~WH Lau, Wanli Ouyang, and Wangmeng Zuo.
\newblock Clip2point: Transfer clip to point cloud classification with image-depth pre-training.
\newblock In \emph{Proceedings of the IEEE/CVF International Conference on Computer Vision}, pages 22157--22167, 2023.

\bibitem[Jiang et~al.(2018)Jiang, Grigorev, Rho, Tian, Fu, Jifara, Adil, and Liu]{medical1}
Feng Jiang, Aleksei Grigorev, Seungmin Rho, Zhihong Tian, YunSheng Fu, Worku Jifara, Khan Adil, and Shaohui Liu.
\newblock Medical image semantic segmentation based on deep learning.
\newblock \emph{Neural Computing and Applications}, 29:\penalty0 1257--1265, 2018.

\bibitem[Kerr et~al.(2023)Kerr, Kim, Goldberg, Kanazawa, and Tancik]{lerf2023}
Justin Kerr, Chung~Min Kim, Ken Goldberg, Angjoo Kanazawa, and Matthew Tancik.
\newblock Lerf: Language embedded radiance fields.
\newblock In \emph{International Conference on Computer Vision (ICCV)}, 2023.

\bibitem[Kobayashi et~al.(2022)Kobayashi, Matsumoto, and Sitzmann]{kobayashi2022distilledfeaturefields}
Sosuke Kobayashi, Eiichi Matsumoto, and Vincent Sitzmann.
\newblock Decomposing nerf for editing via feature field distillation.
\newblock In \emph{Advances in Neural Information Processing Systems}, 2022.

\bibitem[Kundu et~al.(2022)Kundu, Genova, Yin, Fathi, Pantofaru, Guibas, Tagliasacchi, Dellaert, and Funkhouser]{kundu2022panoptic}
Abhijit Kundu, Kyle Genova, Xiaoqi Yin, Alireza Fathi, Caroline Pantofaru, Leonidas~J Guibas, Andrea Tagliasacchi, Frank Dellaert, and Thomas Funkhouser.
\newblock Panoptic neural fields: A semantic object-aware neural scene representation.
\newblock In \emph{Proceedings of the IEEE/CVF Conference on Computer Vision and Pattern Recognition}, pages 12871--12881, 2022.

\bibitem[Lee et~al.(2017)Lee, Zung, Li, Jain, and Seung]{3dresunet}
Kisuk Lee, Jonathan Zung, Peter Li, Viren Jain, and H~Sebastian Seung.
\newblock Superhuman accuracy on the snemi3d connectomics challenge.
\newblock \emph{arXiv preprint arXiv:1706.00120}, 2017.

\bibitem[Li et~al.(2018)Li, Liu, Xu, and Qiu]{driving1}
Baojun Li, Shun Liu, Weichao Xu, and Wei Qiu.
\newblock Real-time object detection and semantic segmentation for autonomous driving.
\newblock In \emph{MIPPR 2017: Automatic Target Recognition and Navigation}, pages 167--174. SPIE, 2018.

\bibitem[Li et~al.(2022)Li, Weinberger, Belongie, Koltun, and Ranftl]{lseg}
Boyi Li, Kilian~Q Weinberger, Serge Belongie, Vladlen Koltun, and Rene Ranftl.
\newblock Language-driven semantic segmentation.
\newblock In \emph{International Conference on Learning Representations}, 2022.

\bibitem[Liang et~al.(2023)Liang, Wu, Dai, Li, Zhao, Zhang, Zhang, Vajda, and Marculescu]{ovseg}
Feng Liang, Bichen Wu, Xiaoliang Dai, Kunpeng Li, Yinan Zhao, Hang Zhang, Peizhao Zhang, Peter Vajda, and Diana Marculescu.
\newblock Open-vocabulary semantic segmentation with mask-adapted clip.
\newblock In \emph{Proceedings of the IEEE/CVF Conference on Computer Vision and Pattern Recognition}, pages 7061--7070, 2023.

\bibitem[Liu et~al.(2023)Liu, Zhang, Zheng, and Duan]{semanticray}
Fangfu Liu, Chubin Zhang, Yu Zheng, and Yueqi Duan.
\newblock Semantic ray: Learning a generalizable semantic field with cross-reprojection attention.
\newblock In \emph{Proceedings of the IEEE/CVF Conference on Computer Vision and Pattern Recognition}, pages 17386--17396, 2023.

\bibitem[Liu et~al.(2022)Liu, Peng, Liu, Wang, Wang, Theobalt, Zhou, and Wang]{neuray}
Yuan Liu, Sida Peng, Lingjie Liu, Qianqian Wang, Peng Wang, Christian Theobalt, Xiaowei Zhou, and Wenping Wang.
\newblock Neural rays for occlusion-aware image-based rendering.
\newblock In \emph{Proceedings of the IEEE/CVF Conference on Computer Vision and Pattern Recognition}, pages 7824--7833, 2022.

\bibitem[Mehta et~al.(2018)Mehta, Rastegari, Caspi, Shapiro, and Hajishirzi]{2d4}
Sachin Mehta, Mohammad Rastegari, Anat Caspi, Linda Shapiro, and Hannaneh Hajishirzi.
\newblock Espnet: Efficient spatial pyramid of dilated convolutions for semantic segmentation.
\newblock In \emph{Proceedings of the european conference on computer vision (ECCV)}, pages 552--568, 2018.

\bibitem[Mildenhall et~al.(2020)Mildenhall, Srinivasan, Tancik, Barron, Ramamoorthi, and Ng]{nerf}
Ben Mildenhall, Pratul Srinivasan, Matthew Tancik, Jonathan Barron, Ravi Ramamoorthi, and Ren Ng.
\newblock Nerf: Representing scenes as neural radiance fields for view synthesis.
\newblock In \emph{European Conference on Computer Vision}, pages 405--421. Springer, 2020.

\bibitem[M{\"u}ller et~al.(2022)M{\"u}ller, Evans, Schied, and Keller]{instant}
Thomas M{\"u}ller, Alex Evans, Christoph Schied, and Alexander Keller.
\newblock Instant neural graphics primitives with a multiresolution hash encoding.
\newblock \emph{ACM Transactions on Graphics (ToG)}, 41\penalty0 (4):\penalty0 1--15, 2022.

\bibitem[Peng et~al.(2023)Peng, Genova, Jiang, Tagliasacchi, Pollefeys, Funkhouser, et~al.]{openscene}
Songyou Peng, Kyle Genova, Chiyu Jiang, Andrea Tagliasacchi, Marc Pollefeys, Thomas Funkhouser, et~al.
\newblock Openscene: 3d scene understanding with open vocabularies.
\newblock In \emph{Proceedings of the IEEE/CVF Conference on Computer Vision and Pattern Recognition}, pages 815--824, 2023.

\bibitem[Pham et~al.(2019)Pham, Hua, Nguyen, and Yeung]{3d1}
Quang-Hieu Pham, Binh-Son Hua, Thanh Nguyen, and Sai-Kit Yeung.
\newblock Real-time progressive 3d semantic segmentation for indoor scenes.
\newblock In \emph{2019 IEEE Winter Conference on Applications of Computer Vision (WACV)}, pages 1089--1098. IEEE, 2019.

\bibitem[Qi et~al.(2017{\natexlab{a}})Qi, Su, Mo, and Guibas]{pointnet}
Charles~R Qi, Hao Su, Kaichun Mo, and Leonidas~J Guibas.
\newblock Pointnet: Deep learning on point sets for 3d classification and segmentation.
\newblock In \emph{Proceedings of the IEEE conference on computer vision and pattern recognition}, pages 652--660, 2017{\natexlab{a}}.

\bibitem[Qi et~al.(2017{\natexlab{b}})Qi, Yi, Su, and Guibas]{pointnet++}
Charles~Ruizhongtai Qi, Li Yi, Hao Su, and Leonidas~J Guibas.
\newblock Pointnet++: Deep hierarchical feature learning on point sets in a metric space.
\newblock \emph{Advances in neural information processing systems}, 30, 2017{\natexlab{b}}.

\bibitem[Radford et~al.(2021)Radford, Kim, Hallacy, Ramesh, Goh, Agarwal, Sastry, Askell, Mishkin, Clark, et~al.]{clip}
Alec Radford, Jong~Wook Kim, Chris Hallacy, Aditya Ramesh, Gabriel Goh, Sandhini Agarwal, Girish Sastry, Amanda Askell, Pamela Mishkin, Jack Clark, et~al.
\newblock Learning transferable visual models from natural language supervision.
\newblock In \emph{International conference on machine learning}, pages 8748--8763. PMLR, 2021.

\bibitem[Schuhmann et~al.(2022)Schuhmann, Beaumont, Vencu, Gordon, Wightman, Cherti, Coombes, Katta, Mullis, Wortsman, et~al.]{laion}
Christoph Schuhmann, Romain Beaumont, Richard Vencu, Cade Gordon, Ross Wightman, Mehdi Cherti, Theo Coombes, Aarush Katta, Clayton Mullis, Mitchell Wortsman, et~al.
\newblock Laion-5b: An open large-scale dataset for training next generation image-text models.
\newblock \emph{Advances in Neural Information Processing Systems}, 35:\penalty0 25278--25294, 2022.

\bibitem[Simonyan and Zisserman(2014)]{2d1}
Karen Simonyan and Andrew Zisserman.
\newblock Very deep convolutional networks for large-scale image recognition.
\newblock \emph{arXiv preprint arXiv:1409.1556}, 2014.

\bibitem[Straub et~al.(2019)Straub, Whelan, Ma, Chen, Wijmans, Green, Engel, Mur-Artal, Ren, Verma, et~al.]{replica}
Julian Straub, Thomas Whelan, Lingni Ma, Yufan Chen, Erik Wijmans, Simon Green, Jakob~J Engel, Raul Mur-Artal, Carl Ren, Shobhit Verma, et~al.
\newblock The replica dataset: A digital replica of indoor spaces.
\newblock \emph{arXiv preprint arXiv:1906.05797}, 2019.

\bibitem[Sun et~al.(2021)Sun, Xie, Chen, Zhou, and Bao]{sun2021neucon}
Jiaming Sun, Yiming Xie, Linghao Chen, Xiaowei Zhou, and Hujun Bao.
\newblock {NeuralRecon}: Real-time coherent {3D} reconstruction from monocular video.
\newblock \emph{CVPR}, 2021.

\bibitem[Tsagkas et~al.(2023)Tsagkas, Mac~Aodha, and Lu]{vl-fields}
Nikolaos Tsagkas, Oisin Mac~Aodha, and Chris~Xiaoxuan Lu.
\newblock Vl-fields: Towards language-grounded neural implicit spatial representations.
\newblock \emph{arXiv preprint arXiv:2305.12427}, 2023.

\bibitem[Tschernezki et~al.(2022)Tschernezki, Laina, Larlus, and Vedaldi]{tschernezki22neural}
Vadim Tschernezki, Iro Laina, Diane Larlus, and Andrea Vedaldi.
\newblock Neural feature fusion fields: {3D} distillation of self-supervised {2D} image representations.
\newblock In \emph{Proceedings of the International Conference on {3D} Vision (3DV)}, 2022.

\bibitem[Tseng and Jan(2018)]{driving2}
Yu-Ho Tseng and Shau-Shiun Jan.
\newblock Combination of computer vision detection and segmentation for autonomous driving.
\newblock In \emph{2018 IEEE/ION Position, Location and Navigation Symposium (PLANS)}, pages 1047--1052. IEEE, 2018.

\bibitem[Vaswani et~al.(2017)Vaswani, Shazeer, Parmar, Uszkoreit, Jones, Gomez, Kaiser, and Polosukhin]{attention}
Ashish Vaswani, Noam Shazeer, Niki Parmar, Jakob Uszkoreit, Llion Jones, Aidan~N Gomez, {\L}ukasz Kaiser, and Illia Polosukhin.
\newblock Attention is all you need.
\newblock \emph{Advances in neural information processing systems}, 30, 2017.

\bibitem[Wang et~al.(2021{\natexlab{a}})Wang, Liu, Liu, Theobalt, Komura, and Wang]{neus}
Peng Wang, Lingjie Liu, Yuan Liu, Christian Theobalt, Taku Komura, and Wenping Wang.
\newblock Neus: Learning neural implicit surfaces by volume rendering for multi-view reconstruction.
\newblock \emph{arXiv preprint arXiv:2106.10689}, 2021{\natexlab{a}}.

\bibitem[Wang et~al.(2022)Wang, Yang, Men, Lin, Bai, Li, Ma, Zhou, Zhou, and Yang]{ofa}
Peng Wang, An Yang, Rui Men, Junyang Lin, Shuai Bai, Zhikang Li, Jianxin Ma, Chang Zhou, Jingren Zhou, and Hongxia Yang.
\newblock Ofa: Unifying architectures, tasks, and modalities through a simple sequence-to-sequence learning framework.
\newblock In \emph{International Conference on Machine Learning}, pages 23318--23340. PMLR, 2022.

\bibitem[Wang et~al.(2021{\natexlab{b}})Wang, Wang, Genova, Srinivasan, Zhou, Barron, Martin-Brualla, Snavely, and Funkhouser]{wang2021ibrnet}
Qianqian Wang, Zhicheng Wang, Kyle Genova, Pratul Srinivasan, Howard Zhou, Jonathan~T. Barron, Ricardo Martin-Brualla, Noah Snavely, and Thomas Funkhouser.
\newblock Ibrnet: Learning multi-view image-based rendering.
\newblock In \emph{CVPR}, 2021{\natexlab{b}}.

\bibitem[Wang et~al.(2004)Wang, Bovik, Sheikh, and Simoncelli]{ssim}
Zhou Wang, Alan~C Bovik, Hamid~R Sheikh, and Eero~P Simoncelli.
\newblock Image quality assessment: from error visibility to structural similarity.
\newblock \emph{IEEE transactions on image processing}, 13\penalty0 (4):\penalty0 600--612, 2004.

\bibitem[Xie et~al.(2017)Xie, Girshick, Doll{\'a}r, Tu, and He]{2d3}
Saining Xie, Ross Girshick, Piotr Doll{\'a}r, Zhuowen Tu, and Kaiming He.
\newblock Aggregated residual transformations for deep neural networks.
\newblock In \emph{Proceedings of the IEEE conference on computer vision and pattern recognition}, pages 1492--1500, 2017.

\bibitem[Xu et~al.(2023)Xu, Wu, Hou, Tsai, Li, Wang, Zhan, He, Vajda, Keutzer, et~al.]{nerfdet}
Chenfeng Xu, Bichen Wu, Ji Hou, Sam Tsai, Ruilong Li, Jialiang Wang, Wei Zhan, Zijian He, Peter Vajda, Kurt Keutzer, et~al.
\newblock Nerf-det: Learning geometry-aware volumetric representation for multi-view 3d object detection.
\newblock In \emph{Proceedings of the IEEE/CVF International Conference on Computer Vision}, pages 23320--23330, 2023.

\bibitem[Xu et~al.(2022)Xu, Xu, Philip, Bi, Shu, Sunkavalli, and Neumann]{xu2022point}
Qiangeng Xu, Zexiang Xu, Julien Philip, Sai Bi, Zhixin Shu, Kalyan Sunkavalli, and Ulrich Neumann.
\newblock Point-nerf: Point-based neural radiance fields.
\newblock \emph{arXiv preprint arXiv:2201.08845}, 2022.

\bibitem[Yang et~al.(2021)Yang, Zhang, Xu, Li, Zhou, Bao, Zhang, and Cui]{yang2021learning}
Bangbang Yang, Yinda Zhang, Yinghao Xu, Yijin Li, Han Zhou, Hujun Bao, Guofeng Zhang, and Zhaopeng Cui.
\newblock Learning object-compositional neural radiance field for editable scene rendering.
\newblock In \emph{Proceedings of the IEEE/CVF International Conference on Computer Vision}, pages 13779--13788, 2021.

\bibitem[Yao et~al.(2021)Yao, Huang, Hou, Lu, Niu, Xu, Liang, Li, Jiang, and Xu]{filip}
Lewei Yao, Runhui Huang, Lu Hou, Guansong Lu, Minzhe Niu, Hang Xu, Xiaodan Liang, Zhenguo Li, Xin Jiang, and Chunjing Xu.
\newblock Filip: Fine-grained interactive language-image pre-training.
\newblock \emph{arXiv preprint arXiv:2111.07783}, 2021.

\bibitem[Yu et~al.(2021)Yu, Ye, Tancik, and Kanazawa]{yu2020pixelnerf}
Alex Yu, Vickie Ye, Matthew Tancik, and Angjoo Kanazawa.
\newblock {pixelNeRF}: Neural radiance fields from one or few images.
\newblock In \emph{CVPR}, 2021.

\bibitem[Yu et~al.(2022)Yu, Wang, Vasudevan, Yeung, Seyedhosseini, and Wu]{coca}
Jiahui Yu, Zirui Wang, Vijay Vasudevan, Legg Yeung, Mojtaba Seyedhosseini, and Yonghui Wu.
\newblock Coca: Contrastive captioners are image-text foundation models.
\newblock \emph{arXiv preprint arXiv:2205.01917}, 2022.

\bibitem[Zhang et~al.(2023)Zhang, Li, and Ahuja]{open-nerf}
Hao Zhang, Fang Li, and Narendra Ahuja.
\newblock Open-nerf: Towards open vocabulary nerf decomposition.
\newblock \emph{arXiv preprint arXiv:2310.16383}, 2023.

\bibitem[Zhang et~al.(2018{\natexlab{a}})Zhang, Isola, Efros, Shechtman, and Wang]{lpips}
Richard Zhang, Phillip Isola, Alexei~A Efros, Eli Shechtman, and Oliver Wang.
\newblock The unreasonable effectiveness of deep features as a perceptual metric.
\newblock In \emph{Proceedings of the IEEE conference on computer vision and pattern recognition}, pages 586--595, 2018{\natexlab{a}}.

\bibitem[Zhang et~al.(2018{\natexlab{b}})Zhang, Chen, He, Ye, Cai, and Zhang]{navi1}
Yuxiao Zhang, Haiqiang Chen, Yiran He, Mao Ye, Xi Cai, and Dan Zhang.
\newblock Road segmentation for all-day outdoor robot navigation.
\newblock \emph{Neurocomputing}, 314:\penalty0 316--325, 2018{\natexlab{b}}.

\bibitem[Zhi et~al.(2021)Zhi, Laidlow, Leutenegger, and Davison]{semantic_nerf}
Shuaifeng Zhi, Tristan Laidlow, Stefan Leutenegger, and Andrew~J. Davison.
\newblock In-place scene labelling and understanding with implicit scene representation.
\newblock In \emph{ICCV}, 2021.

\bibitem[Zhou et~al.(2022)Zhou, Loy, and Dai]{maskclip}
Chong Zhou, Chen~Change Loy, and Bo Dai.
\newblock Extract free dense labels from clip.
\newblock In \emph{European Conference on Computer Vision}, pages 696--712. Springer, 2022.

\bibitem[Zhu et~al.(2022)Zhu, Zhang, He, Zeng, Zhang, and Gao]{pointclip}
Xiangyang Zhu, Renrui Zhang, Bowei He, Ziyao Zeng, Shanghang Zhang, and Peng Gao.
\newblock Pointclip v2: Adapting clip for powerful 3d open-world learning.
\newblock \emph{arXiv preprint arXiv:2211.11682}, 2022.

\end{thebibliography}
}


\end{document}